\documentclass[journal]{IEEEtran}
\usepackage[colorlinks,linkcolor=black,anchorcolor=black,citecolor=black]{hyperref}
\usepackage{graphicx}
\usepackage{dsfont}
\usepackage{stfloats}
\usepackage{subfigure}
\usepackage{amsmath}
\usepackage{cite}
\usepackage{booktabs}
\usepackage{algorithm}
\usepackage{algorithmic}
\usepackage{indentfirst} 
\usepackage{amssymb}
\usepackage{multirow}
\usepackage{fvextra}
\usepackage{bm}
\usepackage{hyperref}
\begin{document}
%
\title{Stagewise Unsupervised Domain Adaptation with Adversarial Self-Training for Road Segmentation of Remote Sensing Images}

\author{Lefei~Zhang,~\IEEEmembership{Senior Member,~IEEE,}
        Meng~Lan,
        Jing Zhang,~\IEEEmembership{Member,~IEEE,}
        and  Dacheng Tao,~\IEEEmembership{Fellow,~IEEE} 

\thanks{Manuscript received 27th February 2021, revised 19th May 2021 and 17th July 2021, accepted 7th august 2021. This work was supported by the National Natural Science Foundation of China under Grant 62076188, the Science and Technology Major Project of Hubei Province (Next-Generation AI Technologies) under Grant 2019AEA170, and the Fundamental Research Funds for the Central Universities under Grant 2042021kf0196. The numerical calculations in this paper had been supported by the supercomputing system in the Supercomputing Center of Wuhan University. Dr. Jing Zhang is supported by ARC project FL-170100117. This work was done during M. Lan's internship at JD Explore Academy.}
\thanks{L. Zhang and M. Lan are with the School of Computer Science, Institute of Artificial Intelligence, Wuhan University, Wuhan China. (e-mail:\{zhanglefei, menglan\}@whu.edu.cn).}
\thanks{J. Zhang is with the School of Computer Science, in the Faculty of Engineering, at the University of Sydney, Sydney, Australia. (e-mail: jing.zhang1@sydney.edu.au).

D. Tao is with JD Explore Academy, China (e-mail: taodacheng@jd.com)}}

\markboth{IEEE Transactions on Geoscience and Remote Sensing}%
{Shell \MakeLowercase{\textit{et al.}}: Stagewise Unsupervised Domain Adaptation with Adversarial Self-Training for Road Segmentation of Remote Sensing Images}
\maketitle

\begin{abstract}
Road segmentation from remote sensing images is a challenging task with wide ranges of application potentials. Deep neural networks have advanced this field by leveraging the power of large-scale labeled data, which, however, are extremely expensive and time-consuming to acquire. One solution is to use cheap available data to train a model and deploy it to directly process the data from a specific application domain. Nevertheless, the well-known domain shift (DS) issue prevents the trained model from generalizing well on the target domain. In this paper, we propose a novel stagewise domain adaptation model called RoadDA to address the DS issue in this field. In the first stage, RoadDA adapts the target domain features to align with the source ones via generative adversarial networks (GAN) based inter-domain adaptation. Specifically, a feature pyramid fusion module is devised to avoid information loss of long and thin roads and learn discriminative and robust features. Besides, to address the intra-domain discrepancy in the target domain, in the second stage, we propose an adversarial self-training method. We generate the pseudo labels of target domain using the trained generator and divide it to labeled easy split and unlabeled hard split based on the road confidence scores. The features of hard split are adapted to align with the easy ones using adversarial learning and the intra-domain adaptation process is repeated to progressively improve the segmentation performance. Experiment results on two benchmarks demonstrate that RoadDA can efficiently reduce the domain gap and outperforms state-of-the-art methods. The code is available at \url{https://github.com/LANMNG/RoadDA}.
\end{abstract}

\begin{IEEEkeywords}
remote sensing, road segmentation, unsupervised domain adaptation, self-training.
\end{IEEEkeywords}

%
\IEEEpeerreviewmaketitle

\section{Introduction}

\IEEEPARstart{R}{oad} segmentation from remote sensing (RS) images is a crucial research topic in remote sensing field. It aims to separate the road areas from the complex background and assign the right label to each pixel of the whole image. Road segmentation has many important applications, such as vehicle navigation \cite{Unsalan}, urban planning \cite{urban, RCNNUnet}, disaster assistance \cite{disaster} and so on. Various methods have been proposed to effectively address the road segmentation task. However, the unsupervised learning based methods usually meet low accuracy for depending on the predefined features \cite{Unsalan, Yuan}. As the rapid development of deep learning technique, supervised learning based convolutional neural networks have shown excellent ability of feature extraction and have been applied to many remote sensing image processing tasks \cite{cheng2018learning,Plaza,cheng2016survey,zhang2016deep,zhang2019learning}. Especially, with the aid of deep learning, road segmentation of remote sensing images has made great progress \cite{roadnet, Bruzzone}.

\begin{figure}
\centering
\includegraphics[width=0.45\textwidth]{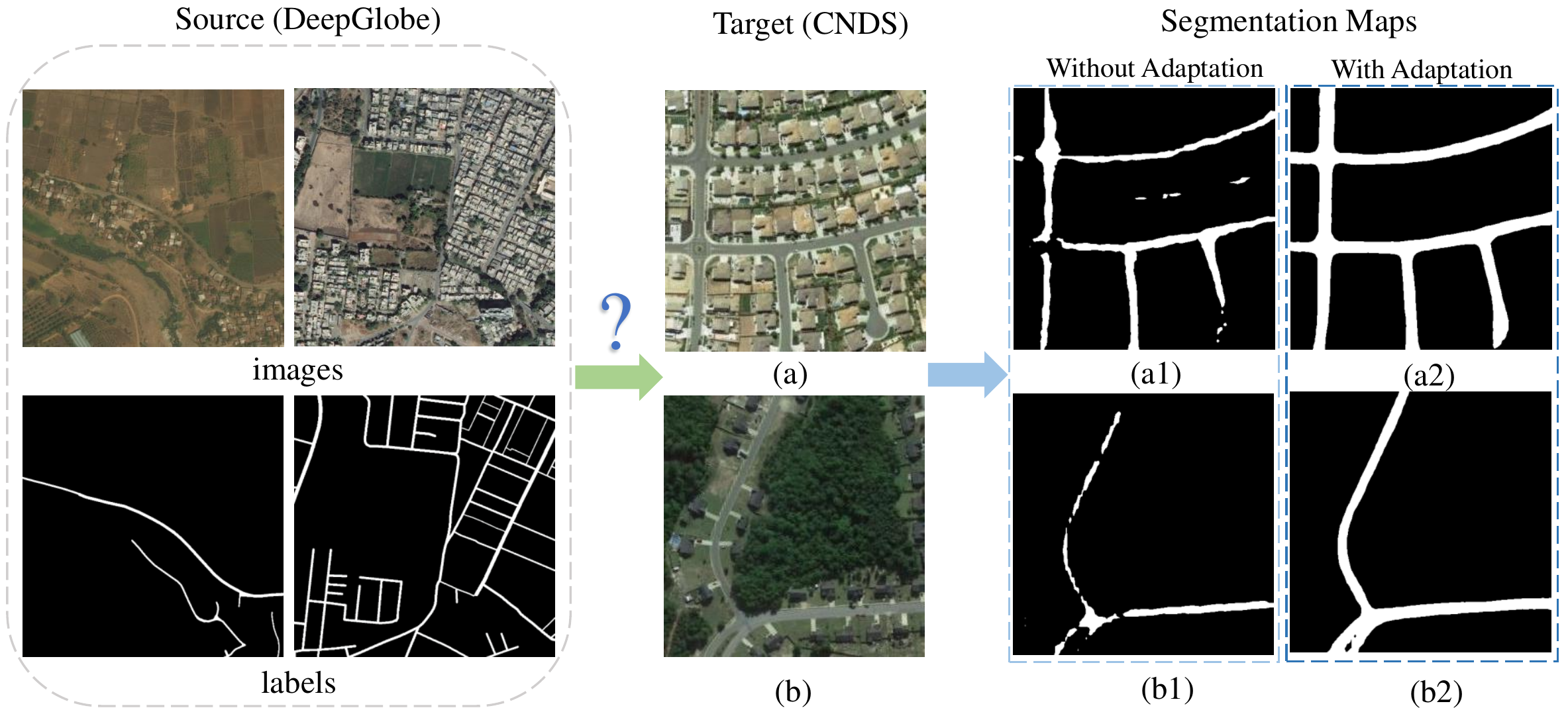}
\caption{How to exploit the available labeled road data (source domain) to improve the segmentation performance of unlabeled road data (target domain) are of practical importance. (a1) and (b1) are the results obtained by directly applying the segmentation model trained on source domain to two target domain images; (a2) and (b2) are the results of our model.}
\label{fig1}
\end{figure}

Many CNNs based road segmentation approaches achive good performance on the public remote sensing benchmark datasets. Cheng et al. \cite{CasNet} proposed a cascaded encoder-decoder network for road segmentation and evaluated it on the collected road dataset which contains 224 manually labeled Google Earth RGB images. \cite{RCNNUnet} designed a RCNN unit with shared weights of convolutional filters to build a deeper network. The model was test on the RoadTracer dataset which obtains the satellite images from Google Map labeled with OpenStreeMap \cite{Tuia}. Zhou et al. \cite{D-LinkNet} introduced the D-LinkNet and won the first prize of Road Extraction Challenge in DeepGlobe 2018, in which the released road dataset has 6226 DigitalGlobe RGB training images. It is exciting to witness various approaches to achieve better performance on different road scenarios, whereas these supervised learning based road segmentation methods always require a large amount of labeled data to train the model and it could be extremely expensive and time-consuming to manually label the road areas from a lot of unlabeled high-resolution remote sensing images in the practical applications.

Fortunately, we have access to some public road datasets released by previous researchers and abundant digital map resources, such as Google Maps and OpenStreetMap, which can be used to create labeled road dataset with minimum efforts. These data may help us to complete the road segmentation task on unlabeled target domain. However, as shown in Fig. \ref{fig1} (a1) and (b1), directly applying the road segmentation model, which is trained on the source domain, to infer the target road images may encounter significant performance drop. It is caused by the domain shift between training images and test images, since these images may have different types of road surfaces (unpaved, paved, dirt roads), rural and urban background areas, etc. Hence, it is important and meaningful to find a suitable technique to address the domain shift issue and take advantage of the available labeled road data to conduct automated road segmentation on target RS images without manual labeling.

Recently, unsupervised domain adaptation (UDA) \cite{colormapgan,zhang2019category} is introduced to solve this problem. Unsupervised domain adaptation utilizes labeled data in one or more relevant source domains to execute new tasks in a target domain with unlabeled data and aims to mitigate the domain shift. Although UDA has been widely studied in remote sensing image classification task \cite{UDA_cla1, crawford}, few efforts have been made in the context of road segmentation. The recent work ASPN \cite{aspn} introduces the random noise to the source domain as synthetic data and performs the adversarial domain adaptation on the output space of source and target domain. However, the existing UDA methods for road segmentation neither fine-tune the segmentation model on the target data with pseudo labels for better performance nor tackle the possible intra-domain discrepancy within the target domain caused during data collection. As a specific technique of semi-supervised learning, self-training \cite{selftraining} could be an useful strategy for UDA. Specifically, it generates pseudo labels of unlabeled target data from the classifier and uses them to fine-tune the model, therefore adapting the model to the target domain. In addition, the adapted model can be used to update the pseudo labels and this procedure can be repeated to improve the performance.

Inspired by these techniques, we propose a novel stagewise unsupervised domain adaptation framework called RoadDA for road segmentation of remote sensing images. RoadDA consists of two stages, namely, the inter-domain adaptation and the adversarial self-training, as illustrated in Fig. \ref{fig:framework}. In the first stage, given the labeled source data and unlabeled target data, a segmentation model serves as the generator to produce the predictions while an inter-domain discriminator predicts the domain labels of these predictions. The target domain are adapted to align with the source domain at the output-level by optimizing the segmentation loss of source domain and the inter-domain adversarial loss. Besides, a feature pyramid fusion module is devised to avoid information loss of long and thin roads and learn discriminative and robust features. In the second stage, to address the intra-domain discrepancy in target domain and further improve the segmentation performance, we design an adversarial self-training scheme. The segmentation model trained in the previous stage is utilized to generate pseudo labels of target domain images. Since road information is the most important prior knowledge in road segmentation task, we pay more attention to the quality of road prediction of the pseudo labels. Therefore, these pseudo labels are checked by a quality estimator according to the confidence score of the road pixels. The retained pseudo labels along with their corresponding images are regarded as the easy split, while the left images in the target domain are regarded as the unlabeled hard split. Similarly, we use the same technique in the first stage to mitigate the intra-domain gap by adapting the distribution of the hard split to that of the easy split. Moreover, we devise a progressive training scheme to iteratively update the pseudo labels from the trained segmentation model and then use them to retrain the segmentation model. During the inference phase, only the adapted segmentation model of the second stage is used for road segmentation without any extra computational requirements.

\par In summary, we conclude our contributions as follows:
\begin{itemize}
    \item [1)] We propose a novel stagewise UDA framework for road segmentation of remote sensing images, which exploits available labeled road data to achieve promising road segmentation on unlabeled target images. This work could be useful in practical application scenarios.
    \item [2)] We devise a new feature pyramid fusion module to avoid the information loss of long and thin roads in high-level feature maps and learn discriminative and robust features for better performance.
    \item [3)] In addition to addressing the inter-domain discrepancy between source and target domains, we propose an adversarial self-training stage to mitigate the intra-domain discrepancy in the target domain. It can progressively achieve intra-domain adaptation and improve road segmentation performance on target domain.
\end{itemize}
\par The rest of the paper is organized as follows. Section \ref{section2} briefly reviews the related work. In Section \ref{section3}, we describe the details of our proposed RoadDA model. Section \ref{section4} provides extensive comparison experiments and ablation studies. Finally, we conclude the paper in Section \ref{section5}.

\section{Related work} \label{section2}

In this section, we review the related methods of road segmentation and unsupervised domain adaptation in the context of remote sensing.

\subsection{Supervised Road Segmentation}
Various applications, such as vehicle navigation and urban planning, require a constantly updated road database, which could be created and updated using road segmentation model on high-resolution remote sensing images. Before the era of deep learning, most of the road segmentation methods are derived from pixel-level classification strategy. Yuan et al. \cite{Yuan} proposed an automatic road extraction approach, where a three-step method including segmentation, medial axis points selection, and road grouping was adopted for road segmentation. Similarly, Das et al. \cite{Das} introduced a multistage algorithm in which probabilistic SVM and salient features were exploited to extract road regions from high-resolution multispectral satellite images.

As the rapid development of deep learning techniques especially the deep convolutional neural networks, they have been widely applied in various kinds of computer vision tasks \cite{Plaza_2,du,zhang2020empowering,wang2020deep,cheng2018deep}, owing to their strong capability of feature learning and end-to-end modeling. Some researchers attempt to introduce deep learning techniques to the field of road segmentation of remote sensing images, which advance the development of this field\cite{CasNet,lan2020,Wei,zhang2019nonlocal}. Cheng et al.\cite{CasNet} designed a cascaded end-to-end CNN for both road segmentation and centerline extraction tasks, where a symmetric encoder-decoder structure was employed for road segmentation. Bruzzone et al. \cite{Bruzzone} proposed a direction-aware residual network for enhancing the structural completeness of the road networks, in which a direction supervision is introduced to strengthen the detection of linear features. Wei et al. \cite{Wei} proposed a multistage framework to accurately extract the road surface and road centerline simultaneously. Li et al. \cite{adv_road} proposed to combine multi-scale global context with adversarial networks. Specifically, the images with ground truth labels and images with prediction maps are fed into the discriminator network, which plays a min-max game with the segmentation model by back-propagating the classification error reversely. Although these approaches achieve remarkable performance, deep supervised learning based methods require a large amount of labeled data, which are expensive and time-consuming to collect and annotate, especially for various unseen scenarios. Since we could get access to some public road datasets released by previous researchers, these data may be useful to help extract road information from unlabeled data by using domain adaptation techniques.

\subsection{Unsupervised Domain Adaptation}

Unsupervised Domain Adaptation (UDA) can be regarded as a particular case of domain adaptation that exploits labeled data in a relevant source domain to conduct similar task in the target domain. The purpose of UDA methodologies is to reduce the domain shift, which typically deteriorates the performance of the models \cite{li2020enhanced,dong2020cscl,song2020unsupervised,chen2019}. In recent years, various UDA methods have been proposed to address the domain shift problem in the semantic segmentation field. Tsai et al. \cite{adaptsegnet} proposed a multi-level adversarial learning framework for UDA semantic segmentation, which employed adversarial learning in the structured output space at different feature levels. Vu et al. Li et al. \cite{li2020model} proposed the collaborative class conditional GAN to bypass the dependence on the source data and improved the performance of the predicted model through generated target-style data. Pan et al. \cite{intraDA} proposed the intra-domain adaptation of the target domain and designed the division criterion for target domain based on the mean of multi-class entropy map. Nevertheless, this criterion may be not suitable for binary road segmentation, since mean entropy based criterion will misclassify all predicted background pixels as easy samples and mislead the intra-domain adaptation. Therefore, we pay more attention to road pixels in the road segmentation task and design a new intra-domain division strategy to better divide the target domain. Furthermore, we propose to combine the intra-domain adaptation with self-training to further improve the segmentation performance.

UDA has also been introduced to the hyperspectral image classification. Tasar et al. \cite{colormapgan} designed the color mapping GAN to generate fake training images with the same semantics as training images and used them to fine-tune the trained classifier. \cite{Ma} used multiple classifiers and variational autoencoders to construct a GAN \cite{gans} to address the domain discrepancy. CaGAN \cite{Xu} proposed a class-aware GAN for UDA multisource remote sensing image classification. CaGAN selects the reliable per-category feature centers based on the clustering and reduces the intra-class and the inter-class discrepancies across domains by optimizing the class-level $l_{1}$-norm loss between the source and target category feature centers. Different from CaGAN, our RoadDA adopts the pseudo labels of easy split and address the intra-class discrepancy within target domain by adversarial self-training. Compared with the multi-class classification task of hyperspectral image, road segmentation can be regarded as a binary classification task of RGB remote sensing images, where more attention should be paid to the road information than background pixels. Moreover, the road shape characteristics and the proportion of road and background pixels should also be considered to improve the model performance. In this sense, UDA methods of hyperspectral image classification that require considerations of spatial-spectral information and interclass relationships may not be suitable for road segmentation. Recently, ASPN \cite{aspn} designed an adversarial spatial pyramid network for the domain adaptation in road segmentation of RS images, which focuses on extracting multi-level effective features and enhancing the feature representation. Few studies have been done on road segmentation of RS images and there are still many issues to be explored and settled.

\begin{figure*}
\centering
\includegraphics[width=0.9\textwidth]{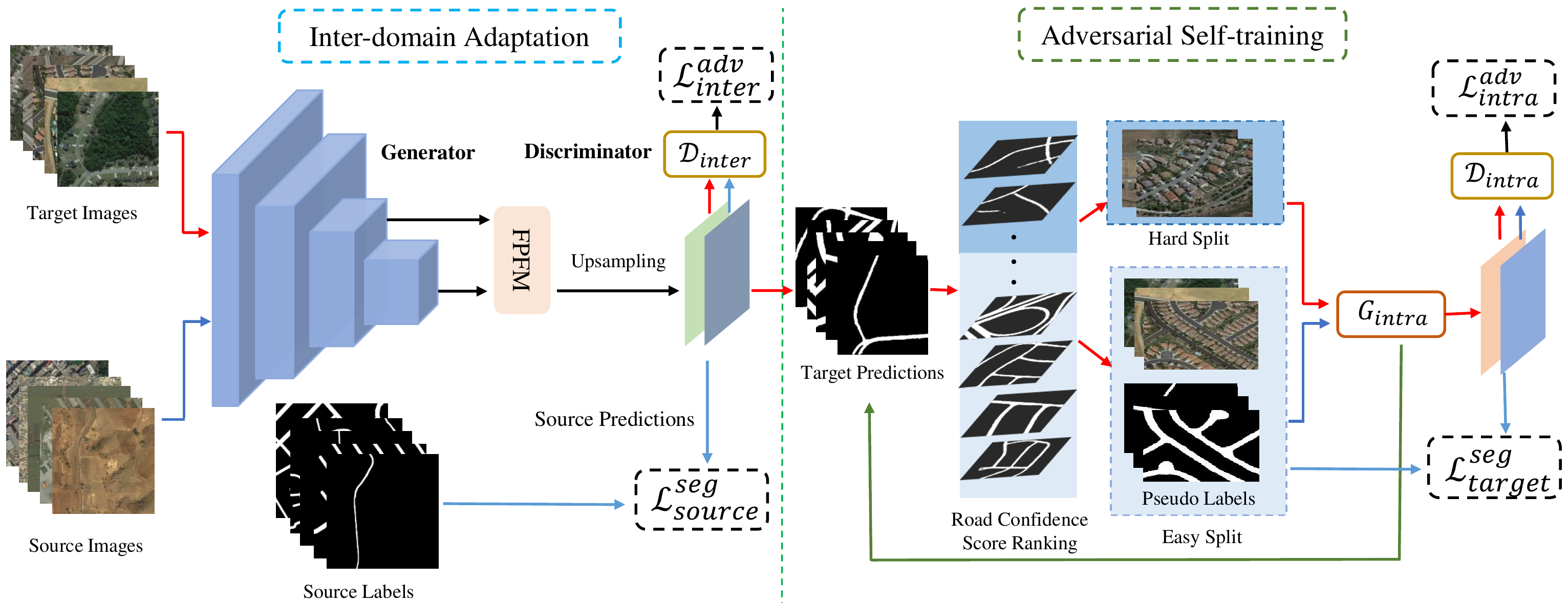}
\caption{The overview of our proposed RoadDA method, which consists of two stages. In the inter-domain adaptation stage, given the source and unlabeled target data, a segmentation model equipped with a specifically devised feature pyramid fusion module (FPFM) acts as the generator to predict the segmentation results. The discriminator is trained to distinguish the domain label of the input while the generator aims to generate similar distribution for both source and target domain to fool the discriminator. In the adversarial self-training stage, we predict the segmentation maps of target domain images using the trained model. The predictions are used to mine the easy target samples by calculating and ranking the road confidence scores and assigned to them as pseudo labels. Theses pseudo labels are used to help adapt the unlabeled hard split to the easy one and therefore reduce the intra-domain discrepancy. The pseudo labels generation and intra-domain adaptation are repeated to gradually improve the performance of the segmentation model.}
\label{fig:framework}
\end{figure*}

\begin{figure}
\centering
\includegraphics[width=0.46\textwidth]{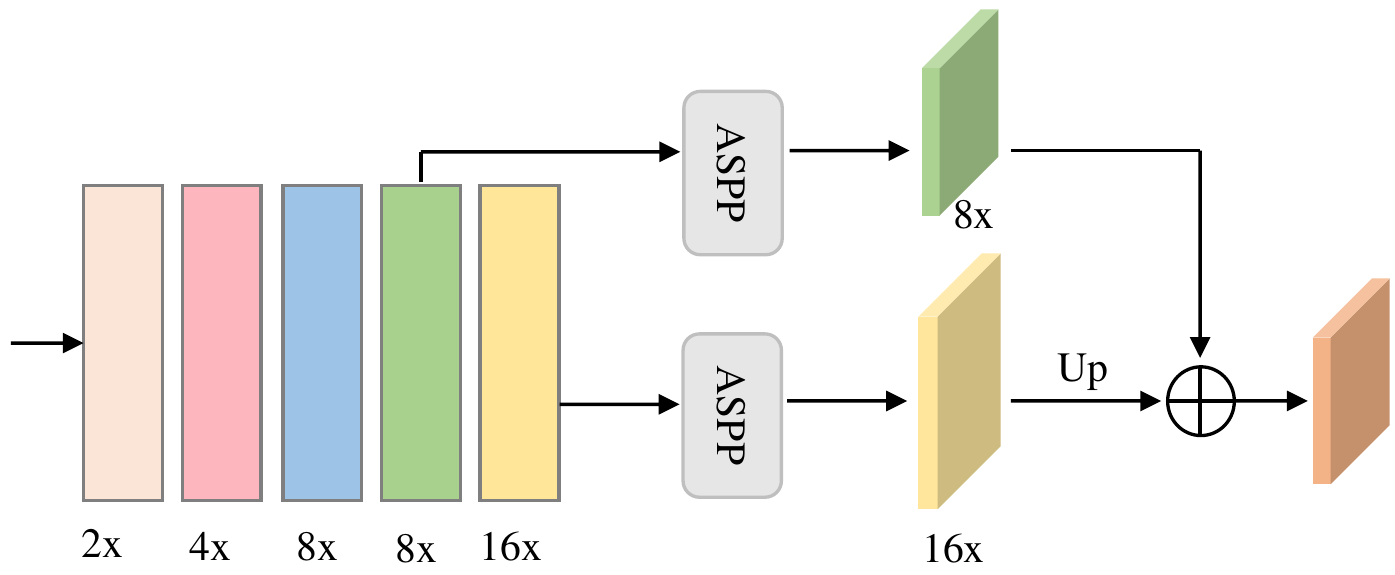}
\caption{Illustration of the proposed feature pyramid fusion module. 8$\times$ denotes a downsampling rate of 8.}
\label{fig_FPFM}
\end{figure}

\section{proposed method}\label{section3}
\subsection{Overview}
Our method aims to improve the performance of road segmentation on unlabeled target data using the unsupervised domain adaptation and self-training techniques. As depicted in Fig. \ref{fig:framework}, RoadDA consists of two stages: inter-domain adaptation stage and adversarial self-training stage. In the first stage, a generative adversarial network (GAN) equipped with a specifically devised feature pyramid fusion module is constructed to reduce the domain shift between the labeled source domain and unlabeled target domain. In the second stage, we use the trained generator to predict the pseudo labels of target domain and then divide the target domain into easy and hard split based on the road confidence scores. An adversarial learning based intra-domain adaptation is performed to fine-tune the trained generator on the target domain and mitigate the intra-domain discrepancy. We adopt the fine-tuned generator to get more accurate pseudo labels and use them for further intra-domain adaptation. This procedure is iterated to progressively improve the segmentation performance until saturation. During the inference phase, only the adaptive generator is used for road segmentation.

\subsection{Inter-domain Adaptation}
Inter-domain adaptation is achieved by the GAN architecture. The input are the labeled source data ($\mathcal{X}_{s}$, $\mathcal{Y}_{s}$) and unlabeled target data ($\mathcal{X}_{t}$), where $\mathcal{X}_{s}, \mathcal{X}_{t} \in \mathbb{R}^{H \times W \times 3}$ and $\mathcal{Y}_{s} \in (0, 1)^{H \times W}$ (0 for background pixel, 1 for road pixel). A segmentation model acts as the generator $G_{inter}$ and predicts the segmentation probability map $P_{s} = G_{inter}(\mathcal{X}_{s})$, $P_{t} = G_{inter}(\mathcal{X}_{t})$ ($P_{s}, P_{t} \in \mathbb{R}^{H \times W \times 2}$) for source images and target images, respectively. As the intermediate representation learned in the segmentation process, the high-level semantic feature of high dimension actually has complex implicit semantics. Therefore, it may be less effective to perform domain alignment at the feature level and there is no guarantee that the joint image-label distributions are aligned between domains. While the binary segmentation predictions of both source images and target images have strong similarity in context and layout. Therefore, we adopt an adversarial learning scheme to align the target domain to the source domain at the output level. We send the predictions $P_{s}$, $P_{t}$ to the discriminator $D_{inter}$ as input to correctly predict their domain labels, while the generator $G_{inter}$ is trained to fool $D_{inter}$, i.e., the generator $G_{inter}$ is encouraged to generate similar prediction distributions on both the target and source domains.

Here, we describe the training process of $G_{inter}$ and $D_{inter}$. For the discriminator, given the segmentation predictions on the source domain and target domain $P_{s}, P_{t}$, the fully convolutional discriminator is trained to classify the domain labels of the input samples. Here we artificially set the labels of source domain samples to 1 and the labels of target domain samples to 0, thus the discriminator could be optimized with the binary cross-entropy (CE) domain classification loss. The original CE loss for discriminator could be formulated as follows:
\begin{equation}
\begin{aligned}
\mathcal{L}_{inter}^{D}\left(P_{s}, P_{t}\right)=&-\sum_{h, w}\left[s \log \left(\mathbf{D}_{inter }\left(P_{s}\right)^{(h, w)}\right)\right.\\
&\left.+(1-s) \log \left(1-\mathbf{D}_{inter}\left(P_{s}\right)^{(h, w)}\right)\right]\\
&-\sum_{h, w}\left[t \log \left(\mathbf{D}_{inter }\left(P_{t}\right)^{(h, w)}\right)\right.\\
&\left.+(1-t) \log \left(1-\mathbf{D}_{inter}\left(P_{t}\right)^{(h, w)}\right)\right].
\end{aligned}
\end{equation}
when s=1 and t=0, it is rewritten as:
\begin{equation}\label{interD_loss}
\begin{aligned}
\mathcal{L}_{inter}^{D}\left(P_{s}, P_{t}\right)=&-\sum_{h, w}\left[\log \left(\mathbf{D}_{inter }\left(P_{s}\right)^{(h, w)}\right)\right.\\
&\left.+\log \left(1-\mathbf{D}_{inter}\left(P_{t}\right)^{(h, w)}\right)\right].
\end{aligned}
\end{equation}

For the training of generator $G_{inter}$, it contains two parts. Firstly, the segmentation loss of images from source domain, which is the cross-entropy loss:
\begin{equation}\label{seg_loss}
\begin{aligned}
\mathcal{L}_{source}^{seg}\left(\mathcal{X}_{s},\mathcal{Y}_{s} \right)=-\sum_{h, w, c} \mathcal{Y}_{s}^{(h, w, c)} \log \left(\mathbf{G}_{inter}\left(\mathcal{X}_{s}\right)^{(h, w, c)}\right),
\end{aligned}
\end{equation}
where the source domain labels $\mathcal{Y}_{s}$ have been converted to the one-hot vector form and $c$ denotes the number of classes. Secondly, the adversarial loss for target images is defined as:
\begin{equation}
\begin{aligned}
\mathcal{L}_{inter}^{adv}\left(\mathcal{X}_{t}\right)=&-\sum_{h, w}\left[t \log \left(\mathbf{D}_{inter}\left(P_{t}\right)^{(h, w)}\right)\right.\\ 
&\left.+(1-t) \log \left(1-\mathbf{D}_{inter}\left(P_{t}\right)^{(h, w)}\right)\right].
\end{aligned}
\end{equation}
According to the adversarial strategy, we set t = 1, which is opposite of the discriminator, then the adversarial loss could be rewritten as:
\begin{equation}\label{inter_adv}
\mathcal{L}_{inter}^{adv}\left(\mathcal{X}_{t}\right)=-\sum_{h, w} \log \left(\mathbf{D}_{inter}\left(\mathbf{G}_{inter}(\mathcal{X}_{t})\right)^{(h, w)}\right).
\end{equation}
This loss forces the generator to produce a source-like distribution on the target domain to fool the discriminator and finally the model reaches a balance state. In summary, the final training objective for the generator is:
\begin{equation}\label{interG_loss}
\mathcal{L}_{inter}^{G}\left(\mathcal{X}_{s}, \mathcal{X}_{t}\right)=\mathcal{L}_{source}^{seg}\left(\mathcal{X}_{s}\right)+\alpha_{adv} \mathcal{L}_{inter}^{adv}\left(\mathcal{X}_{t}\right),
\end{equation}
where $\alpha_{adv}$ is the loss weight to balance the two losses.

During training, We alternatively optimize discriminator $D_{inter}$ and generator $G_{inter}$ using the loss function in Eq.~\eqref{interD_loss} and Eq.~\eqref{interG_loss}, while in inference, only the segmentation model (generator) is used for road segmentation.

\subsubsection{Feature Pyramid Fusion Module}
In the GAN framework, a fully convolutional segmentation model acts as the generator and the backbone network is ResNet-101 \cite{Resnet}. Let the hierarchical features extracted by ResNet-101 be denoted as [$C_{1}, C_{2}, C_{3}, C_{4}, C_{5}$], where the corresponding downsampling rate are [2, 4, 8, 16, 32]. High downsampling rate is originally designed to reduce the feature size and the computation cost and increase the feature robustness. However, in the different scenarios of road segmentation from remote sensing images, high downsampling rate may result in the spatial information loss of long and thin roads, which will affect the final road segmentation performance in dealing with different domain data. In contrast, if we keep high resolution in the deep layer, such as using the downsampling rate set of [2, 4, 8, 8, 8], it will reduce the effective receptive field of the model as well as increase the computation cost in the deep layer. To better handle the various scenarios in both source and target domains, we design the feature pyramid fusion module to fuse multi-level and and multi-scale road features. As shown in Fig. \ref{fig_FPFM}, we employ the downsampling rate set of [2, 4, 8, 8, 16] by setting the stride of fourth stage in ResNet101 to 1, and select the features $C_{4}$ and $C_{5}$ as input to the ASPP module \cite{Deeplab_v2} to generate the global context features $C_{4}^{*}$ and $C_{5}^{*}$ at different level. We then upsample $C_{5}^{*}$ to the same size as $C_{4}^{*}$ and add them together to explicitly enhance the feature representation ability and generate the robust fused feature at the downsampling rate of 8.

\subsection{Adversarial Self-training}
After the training of the inter-domain adaptation stage, we obtain the adapted segmentation model which can achieve considerable performance improvement compared with model trained only on the source domain (source-only). However, according to our observation, there may be intra-domain discrepancy in the target domain due to different illumination conditions and background context. Thus we propose the adversarial learning based self-training stage to further improve the segmentation performance on the target domain. This stage is composed of three parts: intra-domain division, intra-domain adaptation, and self-training. We will describe each part in detail in the following subsection.

\subsubsection{Intra-domain Division}
To address the intra-domain discrepancy in the target domain, it should be decomposed into two parts and reduce their discrepancy. To this end, we propose a quality estimator to accomplish this task. Inspired by the observation that the performance of segmentation model trained on the source domain will deteriorate on the target domain, we use the adapted segmentation model $G_{inter}$ to predict segmentation maps for target domain and accordingly divide the target domain into an easy split and a hard split. Specifically, we design a quality estimator to estimate the road confidence scores of prediction results and rank them based on the scores. Finally, the target images are separated into an easy split $\mathcal{X}_{te}$ with pseudo labels and an unlabeled hard split $\mathcal{X}_{th}$.

Let $\mathcal{P}_{t} = G_{inter}(\mathcal{X}_{t})$ denote the predicted segmentation map of target images, and $\mathcal{P}_{i,j}^{t}$ denotes the probability of road at location $(i,j)$. $\mathcal{M}_{t}$ represents the predicted binary mask (pseudo labels). $\mathcal{M}_{i,j}^{t}$ equals 1 when the probability of pixel at $(i,j)$ is higher than a threshold of 0.5 and the pixel is predicted as road, otherwise it equals 0. Since we focus on the road pixels rather than the large numbers of background pixels which may bias the estimation, we filter out the background and calculate the confidence score of the prediction as follows:
\begin{equation}\label{conf_score}
\mathcal{S}_{cf}=\frac{\sum_{i,j} \mathcal{P}_{i,j}^{t} \cdot \mathcal{M}_{i,j}^{t}}{\sum_{i,j} \mathcal{M}_{i, j}^{t}},
\end{equation}
which is the mean value of the confidence of all predicted road pixels in each image. Then we rank the target images in a descending order according to their $\mathcal{S}_{cf}$ scores and employ a hyperparameter $\lambda \in (0,1)$ as a ratio to divide the ranked target images into an easy split $\mathcal{X}_{te}$ with high-confidence pseudo labels $\mathcal{M}_{te}$ and an unlabeled hard split $\mathcal{X}_{th}$, where $\left|X_{te}\right| = \lambda\left|X_{t}\right|$, $\left|X_{th}\right| = (1-\lambda)\left|X_{t}\right|$. In the experiment section, we conducted a hyperparameter analysis experiment to investigate the influence of $\lambda$.

\subsubsection{Intra-domain Adaptation}
In this part, the same GANs architecture as the first stage is constructed to align the hard split with the easy split at the output level and reduce the intra-domain gap. The generator $G_{intra}$ is initialized with the generator adapted in the first stage and takes $\mathcal{X}_{te}$ and $\mathcal{X}_{th}$ as input, and the output are fed into the discriminator $D_{intra}$, which aims to predict the domain labels. $D_{intra}$ is trained to minimize the binary cross-entropy classification loss while $G_{intra}$ is trained to minimize the segmentation loss and the adversarial loss. They are defined as follows:
\begin{equation}\label{intraD_loss}
\begin{aligned}
\mathcal{L}_{intra}^{D}\left(\mathcal{X}_{te}, \mathcal{X}_{th}\right)=&-\sum_{h, w}\left[\log \left(\mathbf{D}_{intra}\left(\mathbf{G}_{intra}(\mathcal{X}_{te})\right)^{(h, w)}\right)\right.\\
&\left.+\log \left(1-\mathbf{D}_{intra}\left(\mathbf{G}_{intra}(\mathcal{X}_{th})\right)^{(h, w)}\right)\right],
\end{aligned}
\end{equation}
\begin{equation}\label{intraG_loss}
\mathcal{L}_{inter}^{G}\left(\mathcal{X}_{te}, \mathcal{X}_{th}\right)=\mathcal{L}_{target}^{seg}\left(\mathcal{X}_{te}\right)+\beta_{adv} \mathcal{L}_{intra}^{adv}\left(\mathcal{X}_{th}\right),
\end{equation}
where $\beta_{adv}$ is the loss weight to balance these two losses and 
\begin{small}
\begin{equation}\label{segT_loss}
\begin{aligned}
\mathcal{L}_{target}^{seg}\left(\mathcal{X}_{te},\mathcal{M}_{te} \right)=-\sum_{h, w, c} \mathcal{M}_{te}^{(h, w, c)} \log \left(\mathbf{G}_{intra}\left(\mathcal{X}_{te}\right)^{(h, w, c)}\right),
\end{aligned}
\end{equation}
\end{small}
\begin{equation}\label{intra_adv}
\mathcal{L}_{intra}^{adv}\left(\mathcal{X}_{th}\right)=-\sum_{h, w} \log \left(\mathbf{D}_{intra}\left(\mathbf{G}_{intra}(\mathcal{X}_{th})\right)^{(h, w)}\right).
\end{equation}
As in the first stage, $D_{intra}$ and $G_{intra}$ are alternatively optimized using the loss functions in Eq.~\eqref{intraD_loss} and Eq.~\eqref{intraG_loss}.

\subsubsection{Self-training}
After the intra-domain adaptation training on the target domain, the adapted segmentation model $G_{intra}$ achieves better performance and predicts more accurate road segmentation maps. Therefore, we update the pseudo labels of target images using this adapted generator and repeat the intra-domain division and adaptation process to progressively improve the performance of the road segmentation model until saturation, as shown in Fig. \ref{fig:framework}. The training algorithm of RoadDA is summarized in Algorithm \ref{algo1}.

\begin{algorithm}
\caption{The Training Algorithm of RoadDA} \label{algo1}
{\bf Stage1: Inter-domain adaptation}\\
\hspace*{0.2cm} {\bf Input:} source domain data: $D_{s} = (\mathcal{X}_{s}, \mathcal{Y}_{s})$;\\
\hspace*{1.3cm} target domain data: $D_{t} = \mathcal{X}_{t}$.\\
\hspace*{0.2cm} {\bf Initialize:} $G_{inter}$ and $D_{inter}$\\
\hspace*{0.2cm} {\bf while} $epoch < epoch_{max}$ {\bf do}\\
\hspace*{0.5cm} $epoch = epoch + 1$\\
\hspace*{0.5cm} {\bf for} $i < iteration_{max}$; i++ {\bf do}\\
\hspace*{0.8cm} Derive $B_{s}$ and $B_{t}$ sampled from $\mathcal{X}_{s}$ and $\mathcal{X}_{t}$\\
\hspace*{0.8cm} {\bf Train} $G_{inter}$ on $B_{s}$ and $B_{t}$ by optimizing $\mathcal{L}_{inter}^{G}$\\
\hspace*{0.8cm} {\bf Train} $D_{inter}$ on $B_{s}$ and $B_{t}$ by optimizing $\mathcal{L}_{inter}^{D}$\\
\hspace*{0.5cm} {\bf end for}\\
\hspace*{0.2cm} {\bf end while}\\
\hspace*{0.2cm} {\bf Output:} Adapted segmentation model \bm {$G_{inter}$}\\
{\bf Stage2: Adversarial self-training}\\
\hspace*{0.2cm} {\bf Input:} target data: $\mathcal{X}_{t}$, adapted $G_{inter}$, hyperparameter $\lambda$.\\
\hspace*{0.2cm} {\bf Initialize:} Initialize $G_{intra}$ using $G_{inter}$; $D_{intra}$\\
\hspace*{0.2cm} {\bf Step1:} generate pseudo labels $\mathcal{M}_{t}$ for $\mathcal{X}_{t}$ using $G_{intra}$\\
\hspace*{0.2cm} {\bf Step2:} divide $\mathcal{X}_{t}$ into 
$\mathcal{X}_{te}$ and $\mathcal{X}_{th}$ s.t. $\left|\mathcal{X}_{te}\right| = \lambda\left|X_{t}\right|$\\
\hspace*{0.2cm} {\bf Step3:} train $G_{intra}$ on $\mathcal{X}_{te}$ and $\mathcal{X}_{th}$ by optimizing $\mathcal{L}_{intra}^{G}$\\
\hspace*{1.3cm} train $D_{intra}$ on $\mathcal{X}_{te}$ and $\mathcal{X}_{th}$ by optimizing $\mathcal{L}_{intra}^{D}$\\
\hspace*{0.2cm} {\bf Step4:} evaluate the segmentation performance of $G_{intra}$\\
\hspace*{0.2cm} {\bf Repeat:} Step1-Step3\\
\hspace*{0.2cm} {\bf Until:} performance saturation of $G_{intra}$\\
\hspace*{0.2cm} {\bf Output:} \bm {$G_{intra}$}
\end{algorithm}

\subsection{Implementation Details}
For the generator, ResNet-101 is pre-trained on ImageNet. The discriminative fused feature from FPFM is directly upsampled to predict the final segmentaion probability map. For the discriminator, we adopt the fully convolutional network similar to \cite{DCGAN}. The network contains five convolution layers with 4 $\times$ 4 kernel and stride of 2. Leaky ReLU with slope of 0.2 follows each convolution layer as activation function except the last layer.

We implemented our model with PyTorch on a single NVIDIA V100 GPU. The generators ($G_{inter}, G_{intra}$) were optimized using the SGD optimizer with the momentum of 0.9 and a weight decay of $10^{-4}$. We used the Adam optimizer to optimize the discriminator with the momentum of 0.9 and 0.99. The initial learning rates are $4 \times 10^{-4}$ and $1 \times 10^{-4}$ for generators and discriminators, respectively, and both of which are decreased using the polynomial decay policy with a power of 0.9. The loss weights $\alpha_{adv}$ and $\beta_{adv}$ are set to 0.1 and 0.01, respectively. We set the batch size to 4 and $\lambda$ to 0.7.

\section{experiments}\label{section4}
In this section, we conduct extensive experiments to demonstrate the effectiveness of RoadDA in terms of both objective evaluation metrics and subjective visual comparisons.

\subsection{Datasets}
\begin{figure}
\centering
\includegraphics[width=0.45\textwidth]{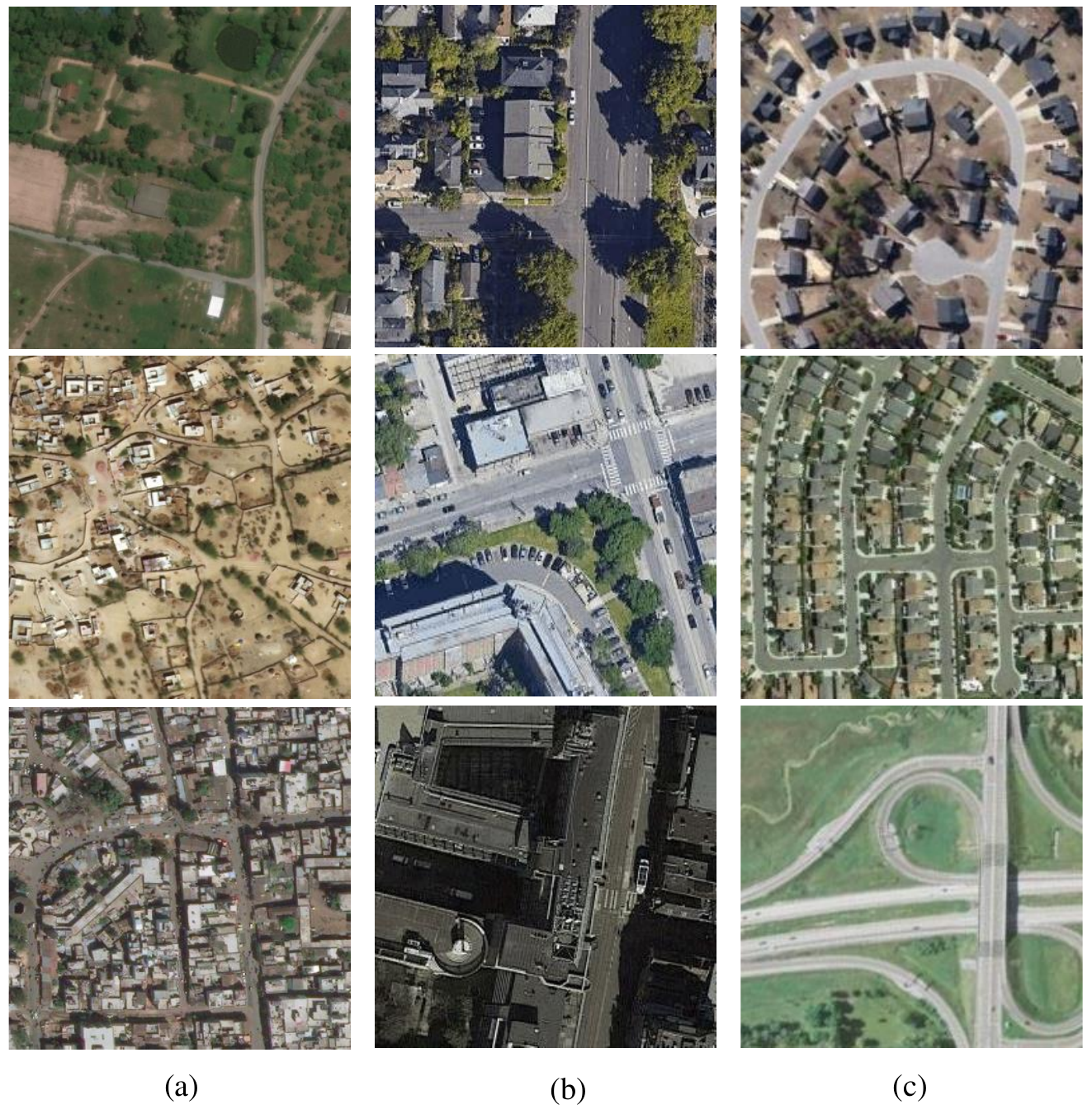}
\caption{Some visual examples from different datasets. (a) DeepGlobe; (b) RTDS; (c) CNDS. }
\label{image_exam}
\end{figure}

In the UDA experiments of road segmentation, we consider the adaptation setting that includes different spatial resolution, background region, and different types of road surfaces. Here, we employ the Roadtracer dataset \cite{RoadCNN} and DeepGlobe dataset \cite{deepglobe} as the source domains, and the CasNet dataset \cite{CasNet} as the target domain. The segmentation model is trained on the labeled source data and unlabeled target data, and is evaluated on the test set of CasNet dataset.
\begin{itemize}
    \item [1)] \textit{DeepGlobe dataset (DeepGlobe)} This dataset is proposed in the road extraction challenge of DeepGlobe 2018 \cite{deepglobe}. The images of this dataset are captured over Thailand, Indonesia, and India with the spatial resolution of 0.5 m/pixel. Images are sampled uniformly between rural and urban areas and cropped to extract useful region by GIS experts. The dataset consists of 8,570 RGB images with size of 1024x1024, where 6,226 images are randomly sampled for training, 1,243 images and 1,101 images are chosen as the validation and test sets, respectively. Specially, the official available dataset only provides labels for the training set. The dataset is available at \url{https://competitions.codalab.org/competitions/18467}.
     \item [2)] \textit{Roadtracer dataset (RTDS)}: RTDS is collected and first used in \cite{RoadCNN}. RTDS is a large corpus of  high-resolutional satellite images and ground-truth road network graphs, in which images are obtained from Google Map with 0.6 m/pixel resolution and ground-truth road network are from the OpenStreeMap (OSM). RTDS covers the urban core of 40 cities in 6 countries. In each city, the center area of approximately 24 $km^{2}$ is selected as the sample of the dataset for a total of 300 RGB images having a size of 4096x4096 pixels. Following \cite{RoadCNN}, images from 25 cities are randomly selected as the training set and images of the 15 remaining cities serve as the test set. The dataset is available at \url{ https://github.com/mitroadmaps/roadtracer}.
    \item [3)] \textit{CasNet dataset (CNDS)}: the CasNet dataset, built by Cheng et al \cite{CasNet}, consists of 224 RGB remote sensing images collected from Google Earth with manual annotations. The images in this dataset are at least 600x600 pixels with the spatial resolution of 1.2 m/pixel. Moreover, these images contain complex background and diversified road shapes. Following \cite{CasNet}, we randomly select 180 images as the training set, 14 images for validation, and the rest 30 images as the test set. The dataset is available at \url{https://shibiaoxu.github.io/TGRS2017.html}.
\end{itemize}
Some image examples are depicted in Fig. \ref{image_exam}.

\subsection{Data Augmentation}
Because of the limitation of the GPU resource. we cannot directly train the model using the high-resolution remote sensing images. Besides, the number of samples in the original dataset is insufficient to train a model with good feature representation and generalization. Therefore, data preprocessing and augmentation are carried out before training.

For each image in these datasets, we randomly crop 20 patches having a size of $512 \times 512$ and then filter out the patches where the number of road pixels is less than 4,000. Since the total number of training samples in CNDS is still insufficient, we flip each cropped training sample in the horizontal direction and rotate the original and flipped samples at the step of $90^{\circ}$ for 3 times. The detailed information of the final processed and augmented datasets is reported in Table \ref{data_info}.

\begin{table}
\renewcommand{\arraystretch}{1.5}
\centering
\caption{Detailed information of the datasets.}
\label{data_info}
\setlength{\tabcolsep}{1.2mm}
\begin{tabular}{|c|c|c|c|c|}
\hline
{Datasets} & Resolution & Size & Training Data& Test Data\\
\hline
DeepGlobe &0.5 m/pixel & $512 \times 512$ & 44,165 & -\\
\hline 
RTDS & 0.6 m/pixel & $512 \times 512$ & 51,106 & - \\
\hline
CNDS & 1.2 m/pixel & $512 \times 512$ & 28,420 & 589 \\
\hline
\end{tabular}
\end{table}

\subsection{Evaluation Metrics}
\par Four common metrics are used to evaluated the quantitative performance of different road segmentation models, including intersection-over-union (IoU), completeness (COM), correctness (COR), and F1 score. COM measures the proportion of matched road pixels in the ground truth map, COR reports the percentage of matched road areas in the predicted segmentation map. The F1 score is a harmonic average between COM and COR which can measure the robustness of methods. The four metrics are defined as follows:
\begin{equation}\label{eqn:RDB_unit}
\begin{aligned}
     &\mathrm{IoU} =\frac{\mathrm{TP}}{\mathrm{TP}+\mathrm{FN}+\mathrm{FP}}
     &\mathrm{COM} =\frac{\mathrm{TP}}{\mathrm{TP}+\mathrm{FN}}\\
     &\mathrm{COR} =\frac{\mathrm{TP}}{\mathrm{TP}+\mathrm{FP}} 
     &\mathrm{F1} =\frac{2 \times \mathrm{COM} \times \mathrm{COR}}{\mathrm{COM}+\mathrm{COR}},
\end{aligned}
\end{equation}
where TP, FP, and FN denote the number of true positive pixels, false positive pixels, and false negative pixels, respectively. A larger metric value indicates better performance.

\subsection{Comparative Methods}
To demonstrate the superiority of the proposed method, we compare it with several state-of-the-art methods, including AdaptSegNet \cite{adaptsegnet}, ADVENT \cite{ADVENT}, BDL \cite{BDL}, and IntraDA \cite{intraDA}. Here, we briefly describe these approaches as follows:
\begin{itemize}
    \item [1)] \textit{Source-only}: This is a baseline model which is only trained on the labeled source domain data while directly tested on the target domain. 
    \item [2)] \textit{Target-only}: Target-only represents the typical supervised road segmentation method which is trained on the target domain with labels. It is the oracle model to provide the upper bound of performance and indicate the effectiveness and reliability of those UDA methods.
    \item [3)] \textit{AdaptSegNet}: AdaptSegNet firstly adopts adversarial learning in the output space for UDA semantic segmentation. Compared with the adaptation in the feature-level, structured output spaces of source and target domain contain the global context and spatial similarities, which is beneficial to the adaptation. Moreover, a multi-level adversarial network is designed to enhance the model.
    \item [4)] \textit{ADVENT}: Based on AdaptSegNet, ADVENT calculates the entropy of output space to avoid low-confident predictions on the target domain. The entropy map is fed into the discriminator for adversarial training.
    \item [5)] \textit{BDL}: BDL adopts a bidirectional learning framework which first carries out image-to-image translation and then uses the translated images to complete the adversarial domain adaptation. The image translation and adversarial adaptation promote each other to progressively improve performance.
    \item [6)] \textit{IntraDA}: In addition to the adaptation between source and target domain, IntraDA tries to address the intra-domain shift, although the entropy based criterion designed to separate the multi-class target domain is not suitable for binary road segmentation tasks. 
\end{itemize}

\subsection {UDA Results on different adaptation setting}
\textbf{DeepGlobe $\longrightarrow$ CNDS.} In Table \ref{dpg_CNDS}, we report the road segmentation performance of our method and other comparison methods on the CNDS test set. For a fair comparison, the baseline model and target-only model adopt the same segmentation model as RoadDA and all the methods also use the same ResNet-101 backbone. Overall, RoadDA achieves the best performance of 74.92\% IoU and 85.81\% F1 score compared with the state-of-the-art UDA methods, which shows the reliability and robustness of our model. Source-only and target-only models produce the worst and best results, respectively, as expected. Compared with the methods with only inter-domain adaptation, such as AdaptSegNet and ADVENT, the intra-domain adaptation and self-training adopted in our model bring considerable performance improvement. For instance, RoadDA is 15.52\% higher than ADVENT in terms of IoU. Moreover, our method also outperforms IntraDA, which also uses intra-domain adaptation. We argue that the performance gain owes to the employment of more effective division criterion for target domain and self-training.

\begin{figure*}
\centering
\includegraphics[width=0.95\textwidth]{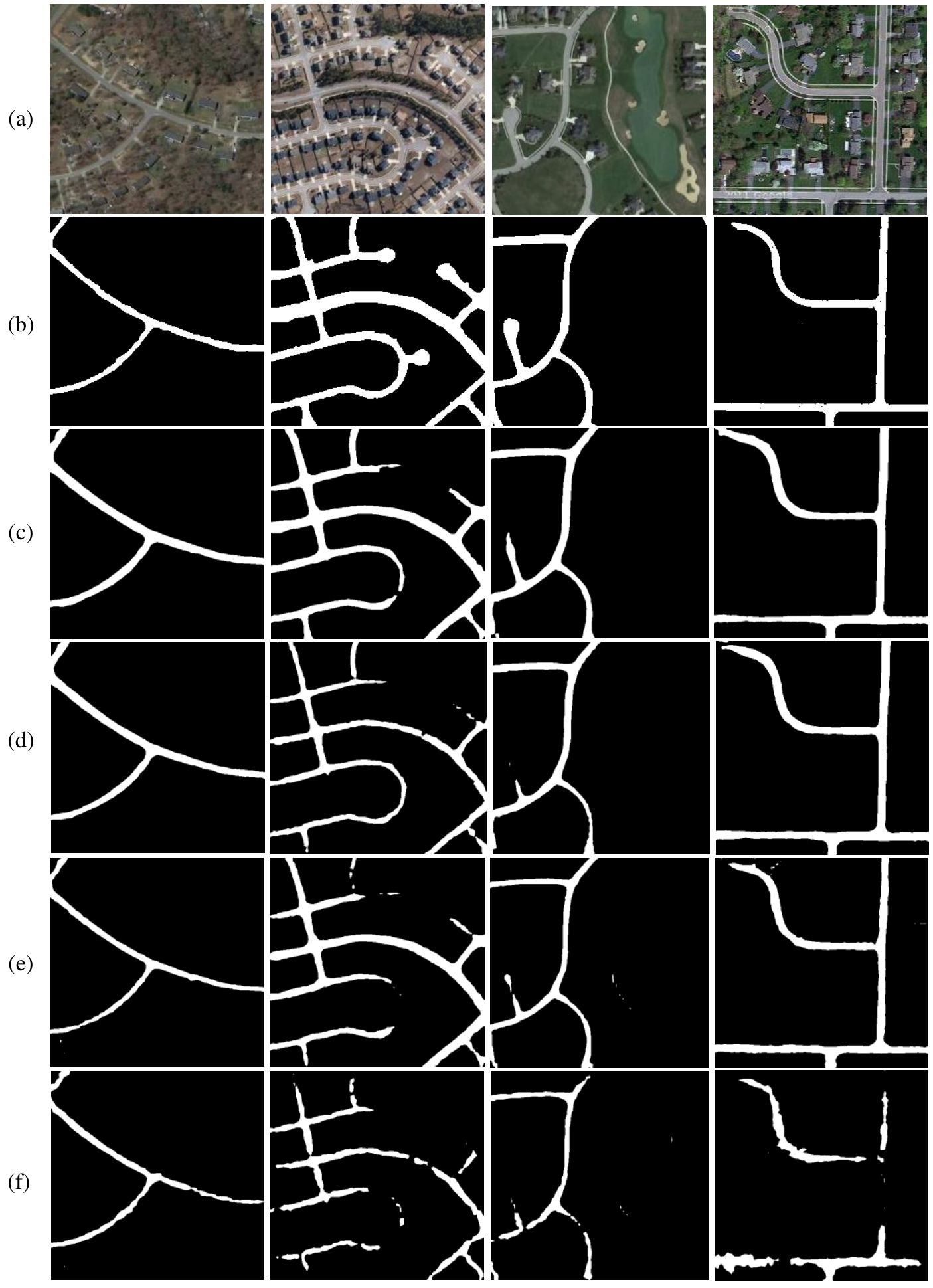}
\caption{Visual results of typical methods on CNDS. (a) the target images. (b) The ground-truth. (c) Our RoadDA. (d) IntraDA. (e) ADVENT. (f) Source-only.}
\label{vis_img}   
\end{figure*}

Fig.\ref{vis_img} presents the visual road segmentation results of different methods on four representative test images from the CNDS test set. It can be observed that RoadDA obtains the best segmentation results close to the ground-truth and could handle the remote sensing images with different complex backgrounds. Moreover, most of the road contours in the images can be captured by RoadDA, demonstrating the great potential of using domain adaptation techniques to solve the unsupervised road segmentation problem.

\textbf{RTDS $\longrightarrow$ CNDS.} Here, we use the RTDS as the source domain and report the evaluation results on the CNDS test set in Table \ref{rtds_CNDS}. All the methods employ the same configuration as the previous experiment. As shown in Table \ref{rtds_CNDS}, our proposed method achieves 61.76\% IoU and 77.48\% F1 score, which is superior to all the UDA comparison methods. Specially, RoadDA is 15.8\% higher than source-only and 7.32\% higher than ADVENT in terms of IoU. Compared with IntraDA equipped with intra-domain adaptation, our model also brings 5.7\% improvement in IoU metric.

Comparing the experimental results of the two adaptation settings, we can find that the adaptation performance of DeepGlobe is better than RTDS. This may be caused by the inherent properties of the domain itself. The images in DeepGlobe dataset have more similar spatial information to the target dataset, such as the background context and road shape, while the images in RTDS are captured in the big city centers where roads are too wide or indistinguishable from surrounding buildings. This may give us some hints for choosing the suitable source domain in practice.

\begin{table*}
\centering
\caption{Quantitative results on the setting of DeepGlobe $\longrightarrow$ CNDS. Adv. learning: adversarial learning.}
\label{dpg_CNDS}
\renewcommand{\arraystretch}{1.4} 
\addtolength{\tabcolsep}{2pt} 
\begin{tabular}{c|ccccccc}
\toprule
\multicolumn{8}{c}{DeepGlobe $\longrightarrow$ CNDS}\\
\toprule
{Methods} & backbone & Input size & Strategy & IoU & COM & COR  & F1 \\ 
\hline
Source-only & ResNet-101 & $512 \times 512$& Segmentation & 52.84 & 57.47 & 89.57 & 70.01\\
\hline
AdaptSegNet \cite{adaptsegnet} & ResNet-101 & $512 \times 512$& Adv. learning & 54.13 & 59.04 & $\mathbf{91.86}$ & 71.88\\
ADVENT \cite{ADVENT} & ResNet-101 & $512 \times 512$& Adv. learning & 59.40 & 72.76 & 90.66 & 80.72\\
BDL \cite{BDL}& ResNet-101 & $512 \times 512$& Style transfer \& Adv. learning & 72.35 & 79.91 & 89.97 & 84.64 \\
IntraDA \cite{intraDA}& ResNet-101 & $512 \times 512$& Adv. learning & 72.73 & 80.06 & 90.21 & 84.83 \\
RoadDA & ResNet-101 & $512 \times 512$& Adv. learning \& self-training & $\mathbf{74.92}$ & $\mathbf{82.11}$ & 89.88 & $\mathbf{85.81}$ \\
\hline
Target-only & ResNet-101 & $512 \times 512$& Segmentation & 85.35 & 91.23 & 93.51 & 92.35\\
\bottomrule
\end{tabular}
\end{table*}

\begin{table*}
\centering
\caption{quantitative results of adapting RTDS to CNDS. Adv. learning: adversarial learning}
\label{rtds_CNDS}
\renewcommand{\arraystretch}{1.4}
\addtolength{\tabcolsep}{2pt}
\begin{tabular}{c|ccccccc}
\toprule
\multicolumn{8}{c}{RTDS $\longrightarrow$ CNDS}\\
\toprule
{Methods} & backbone & Input size & Strategy & IoU & COM & COR  & F1 \\ 
\hline
Source-only & ResNet-101 & $512 \times 512$& Segmentation & 45.96 & 48.39 & $\mathbf{91.36}$ & 63.26\\
\hline
AdaptSegNet \cite{adaptsegnet} & ResNet-101 & $512 \times 512$& Adv. learning & 53.34 & 58.03 & 89.48 & 70.40\\
ADVENT \cite{ADVENT} & ResNet-101 & $512 \times 512$& Adv. learning & 54.44 & 58.03 & 90.71 & 71.26\\
BDL \cite{BDL} & ResNet-101 & $512 \times 512$& Style transfer \& Adv. learning & 55.47 & 59.65 & 90.75 & 71.98 \\
IntraDA \cite{intraDA} & ResNet-101 & $512 \times 512$& Adv. learning & 56.06 & 60.26 & 91.16 & 72.55 \\
RoadDA & ResNet-101 & $512 \times 512$& Adv. learning \& self-training & $\mathbf{61.76}$ & $\mathbf{67.56}$ & 90.81 & $\mathbf{77.48}$ \\
\hline
Target-only & ResNet-101 & $512 \times 512$& Segmentation & 85.35 & 91.23 & 93.51 & 92.35\\
\bottomrule
\end{tabular}
\end{table*}

\subsection{Model Analysis }
\textit{1) Ablation study: } In this part, we evaluate the effects of the proposed modules in RoadDA on the setting of DeepGlobe $\longrightarrow$ CNDS. The input image size is 512 $\times$ 512.

In Table \ref{FPFM}, we summarize the performance of RoadDA using different variants of feature pyramid fusion module on the setting of DeepGlobe $\longrightarrow$ CNDS. RoadDA with 32$\times$ means that we only use the deep feature of a downsampling rate of 32 as input for ASPP module and then directly generate the segmentation map with out the fusion operation. Among different variants of RoadDA only using the single-level feature, the setting of 8$\times$ performs best by obtaining a 74.17\% IoU which shows that the high-resolution features preserve more road spatial information and beneficial to the performance. RoadDA with 32$\times$, 16$\times$, and Fusion denotes using the FPFM which takes the two levels of features of the downsampling rate of 32 and 16 as input. Overall, RoadDA using the proposed setting of FPFM achieves the best performance of 74.92\% IoU compared with all other variants.

We also conducted an ablation study on the second adversarial self-training stage. The results are summarized in Table \ref{adv_selftrining}. As can be seen, RoadDA without intra-domain adaptation and self-training, which only employs the inter-domain adaptation, achieves 61.74\% IoU and 81.13\% F1 score. After using the intra-domain adaptation and self-training strategies, the performance is improved. The intra-domain adaptation provides 11.9\% IoU improvement and the self-training brings 12.34\% IoU gains. RoadDA using both intra-domain adaptation and self-training achieves a gain of 13.18\% IoU over the vanilla baseline, which confirms the effectiveness of the proposed techniques in the adversarial self-training stage.

\textit{2) Division ratio $\lambda$: } For intra-domain division, $\lambda$ controls the number of samples in the easy and hard splits and also has a impact on the distribution of simple samples and difficult samples. To investigate its influence, we conducted the experiments using $\lambda$ from 0.6 to 0.9 on both adaptation settings. As reported in Table \ref{hyperparameter}, RoadDA achieves the best performance for both settings when $\lambda$ = 0.7.

\begin{table}[!htbp]
\centering
\caption{Ablation study results of the proposed FPFM on the setting of DeepGlobe $\longrightarrow$ CNDS.}
\label{FPFM}
\renewcommand{\arraystretch}{1.2}
\addtolength{\tabcolsep}{1.2pt}
\begin{tabular}{cccccc}
\hline
Model & 32$\times$ & 16$\times$ & 8$\times$ & Fusion & IoU\\
\hline
RoadDA & $\checkmark$ & $\times$ & $\times$ & $\times$ & 73.24\\ 
RoadDA & $\times$ & $\checkmark$ & $\times$ & $\times$ & 73.65\\
RoadDA & $\times$ & $\times$ & $\checkmark$ & $\times$ & 74.17\\
RoadDA & $\checkmark$ & $\checkmark$ & $\times$ & $\checkmark$ & 73.98\\
RoadDA & $\times$ & $\checkmark$ & $\checkmark$ & $\checkmark$ & $\mathbf{74.92}$\\
\hline
\end{tabular}
\end{table}

\begin{table}[!htbp]
\centering
\caption{Ablation study results of the proposed techniques in the adversarial self-training stage on the setting of DeepGlobe $\longrightarrow$ CNDS. IA: intra-domain adaptation. ST: self-training.}
\label{adv_selftrining}
\renewcommand{\arraystretch}{1.2}
\addtolength{\tabcolsep}{1.0pt}
\begin{tabular}{ccccccc}
\hline
Model & IA & ST & IoU & COM & COR & F1\\
\hline
RoadDA & $\times$ & $\times$ & 61.74 & 73.48 & 90.57 & 81.13\\
RoadDA & $\checkmark$ & $\times$ & 73.64 & 80.72 & 90.03 & 85.12\\
RoadDA & $\times$ & $\checkmark$ & 74.08 & 81.15 & 89.97 & 85.33\\
RoadDA & $\checkmark$ & $\checkmark$ & $\mathbf{74.92}$ & $\mathbf{82.11}$ & 89.88 & $\mathbf{85.81}$ \\
\hline
\end{tabular}
\end{table}

\begin{table}[!htbp]
\centering
\caption{IoU results of different hyperparameter $\lambda$ on both adaptation settings. DPG: DeepGlobe dataset.}
\label{hyperparameter}
\renewcommand{\arraystretch}{1.5}
\addtolength{\tabcolsep}{2pt}
\begin{tabular}{|c|c|c|c|c|}
\hline
$\lambda$ & 0.6 & 0.7 & 0.8 & 0.9\\
\hline
DPG $\longrightarrow$ CNDS & 73.54 & $\mathbf{74.92}$ & 74.01 & 72.87\\
\hline
RTDS $\longrightarrow$ CNDS & 60.25 & $\mathbf{61.76}$ & 61.33 & 60.62\\
\hline
\end{tabular}
\end{table}

\begin{figure}
\centering
\includegraphics[width=0.45\textwidth]{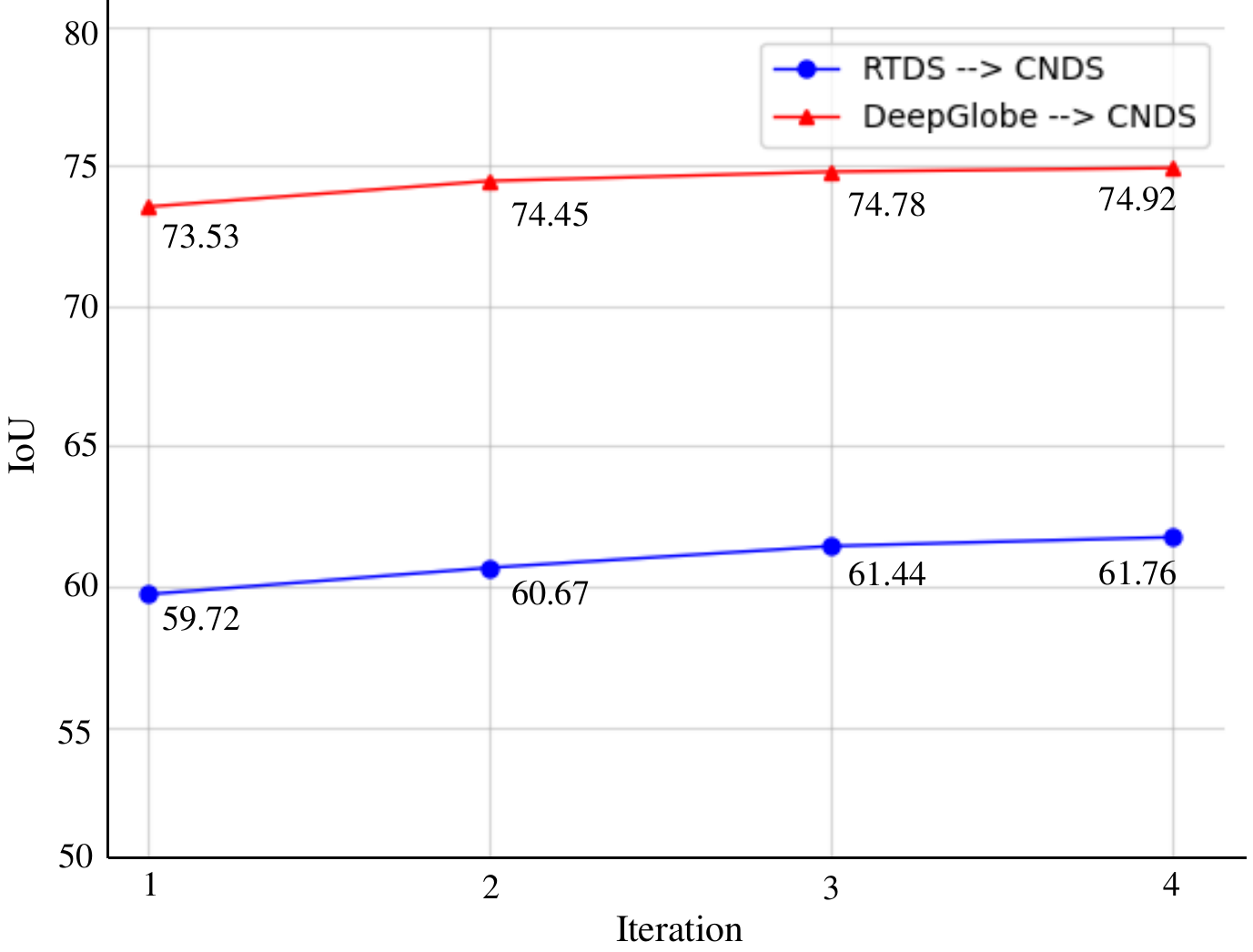}
\caption{IoU results versus the iterations of self-training.}
\label{fig_iter}
\end{figure}

\begin{figure*}[t!]
\centering
\includegraphics[width=0.92\textwidth]{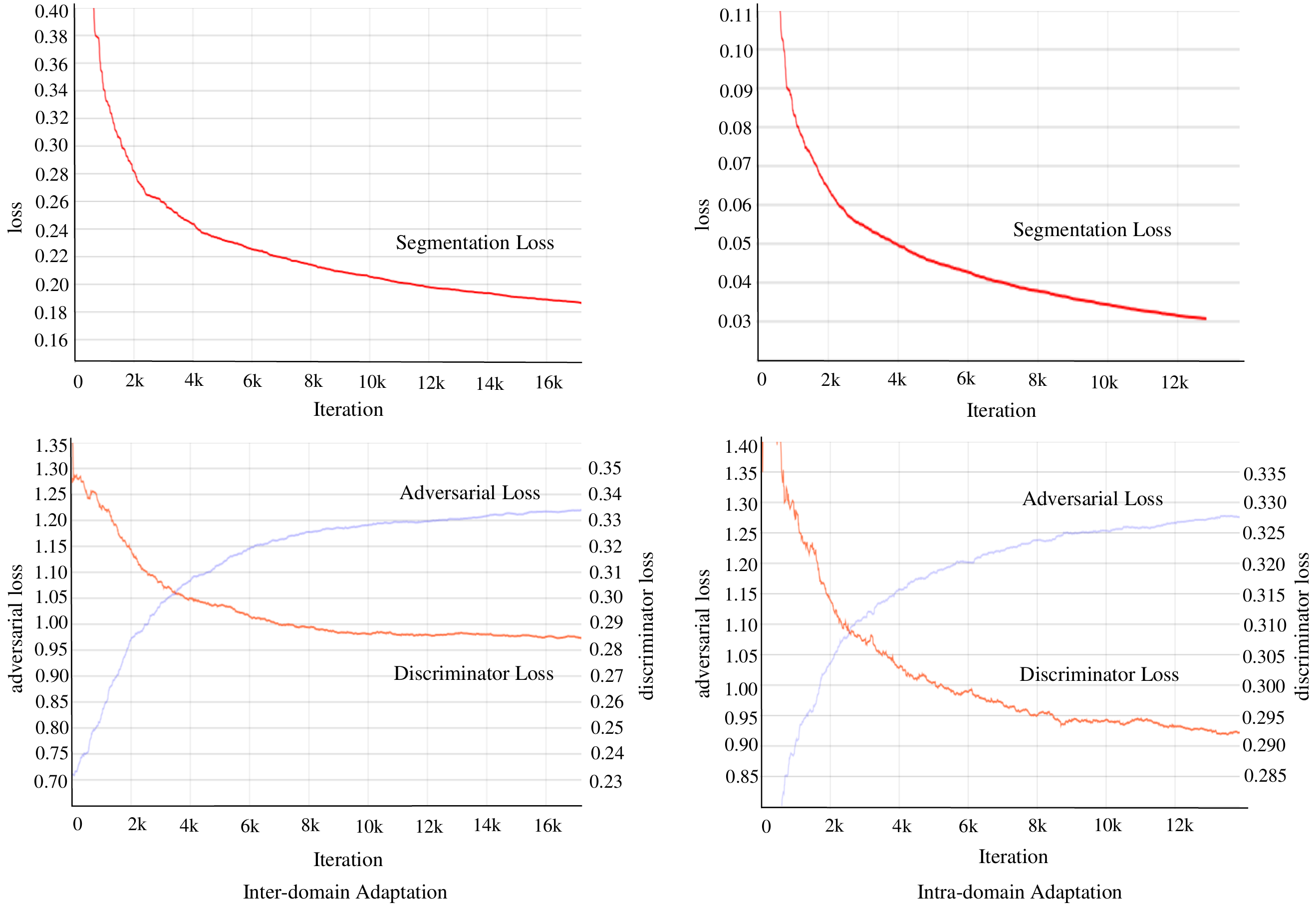}
\caption{Segmentation loss and adversarial loss versus the iterations while training RoadDA on DeepGlobe and CNDS.}
\label{loss_fig}
\end{figure*}

\begin{figure*}[t!]
\centering
\includegraphics[width=0.96\textwidth]{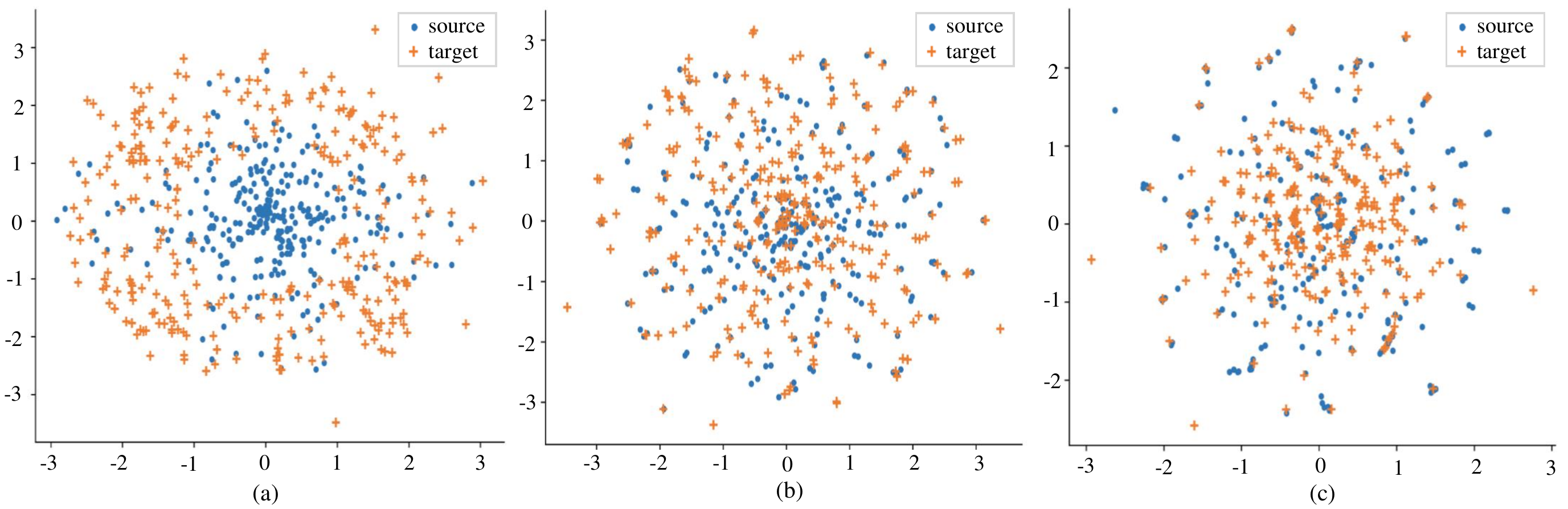}
\caption{T-SNE visualization of the features obtained by different models on the setting of DeepGlobe $\longrightarrow$ CNDS. (a) Source-only. (b) RoadDA using inter-domain adaptation only. (c) RoadDA.}
\label{tsne}
\end{figure*}

\begin{table}[!htbp]
\centering
\caption{The results of two choices of domain adaptation space on the setting of DeepGlobe $\longrightarrow$ CNDS.}
\label{adaptation_space}
\footnotesize
\renewcommand{\arraystretch}{1.5}
\addtolength{\tabcolsep}{1pt}
\begin{tabular}{|c|c|c|c|c|}
\hline
Methods & IoU & COM & COR & F1 \\
\hline
Feature-level adaptation& 51.42 & 56.36 & 87.65 & 68.61 \\
\hline 
RoadDA (output space) & 74.92 & 82.11 & 89.88 & 85.81 \\
\hline
\end{tabular}
\end{table}

\textit{3) Iterarion for self-training: } In the adversarial self-training stage, we iterate the intra-domain adaptation process to progressively improve model performance, while it is unknown how many iterations it will take to achieve performance saturation. Therefore, we analyze the relationship between the iteration and the IoU on two adaptation setting. As depicted in Fig. \ref{fig_iter}, the performance gradually increase and tends to be saturated after three or four iterations. Thus, three or four iterations may be a proper choice to save the training time while achieve a decent performance.

\textit{4) Domain adaptation space:} Here, we investigate the influence of domain adaptation space on the segmentation results. As shown in Table \ref{adaptation_space}, when RoadDA performs the feature-level domain adaptation, the model suffers significant performance deterioration and only achieves 51.42\% IoU which is 23.5\% lower than RoadDA using output space domain adaptation. It indicates that the structured prediction space is more appropriate for domain adaptation in the binary road segmentation task since remote sensing images always contain complex structures and consequently complex feature distributions may obtain for different domains.

\textit{5) Model training: } Fig. \ref{loss_fig} presents the curves of the segmentation loss and adversarial loss against the number of iterations on the DeepGlobe $\longrightarrow$ CNDS setting. The loss curves in the inter-domain adaptation stage and the intra-domain adaptation are listed in the left side and right side, respectively. As can be seen, with the increasing of iterations, the segmentation loss and discriminator loss become smaller until convergence while the adversarial loss increases gradually until convergence. Finally, the adversarial learning reaches a balanced state where the model generates similar distributions on both target and source domains. In addition, we show the t-SNE visualization \cite{tsne} of the features before the prediction layer of two variants of our RoadDA and the source-only model in Fig. \ref{tsne}. The results demonstrate that using the two-stage adaptation in the output-level space, the features in target domain is gradually aligned with the those in the source domain.

\section{Conclusion}\label{section5}
\par In this paper, we propose a novel two-stage unsupervised domain adaptation framework for road segmentation of remote sensing images. In the first inter-domain adaptation stage, our model equipped with a specifically designed feature pyramid fusion module efficiently mitigates the domain discrepancy between the labeled source domain and the unlabeled target domain, resulting in an adapted segmentation model that can generalize well to the target domain. In the second adversarial self-training stage, our model effectively mine easy target samples and assign pseudo labels to them, which are used to guide the intra-domain adaptation to mitigate the discrepancy within the target domain. Our model outperforms state-of-the-art domain adaptation methods on two benchmark settings for remote sensing road segmentation, showing a promising application potential in real-world scenarios. In the future works, we will attempt to use the remote sensing image dataset of river, which has similar shape with road, as the source domain to guide the segmentation of road in target domain.

\bibliography{bibfile}

\vspace{-12mm}
\begin{IEEEbiography}[{\includegraphics[width=0.9in,clip,keepaspectratio]{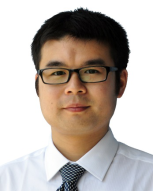}}]{Lefei Zhang (S’11-M’14-SM’20)}
received the B.S. and Ph.D. degrees from Wuhan University, Wuhan, China. He is currently a professor with the School of Computer Science, Wuhan University. He was a Big Data Institute Visitor with the Department of Statistical Science, University College London, and a Hong Kong Scholar with the Department of Computing, The Hong Kong Polytechnic University. His research interests include pattern recognition, image processing, and remote sensing.

Dr. Zhang serves as an associate editor for Pattern Recognition and IEEE Geoscience and Remote Sensing Letters.
\end{IEEEbiography}
\vspace{-12mm}
\begin{IEEEbiography}[{\includegraphics[width=0.9in,clip,keepaspectratio]{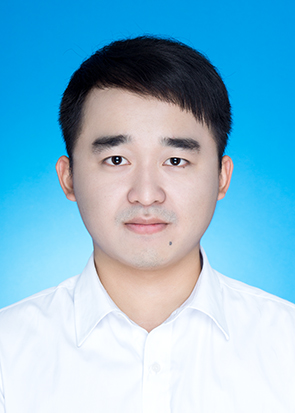}}]{Meng Lan}
 received the B.S. degree from the School of Computer Science, Wuhan University, China, in 2018. He is currently pursuing the Ph.D degree in the School of Computer Science, Wuhan University, China. 
 His research interests include deep learning and computer vision.
\end{IEEEbiography}
\vspace{-12mm}
\begin{IEEEbiography}[{\includegraphics[width=0.9in,clip,keepaspectratio]{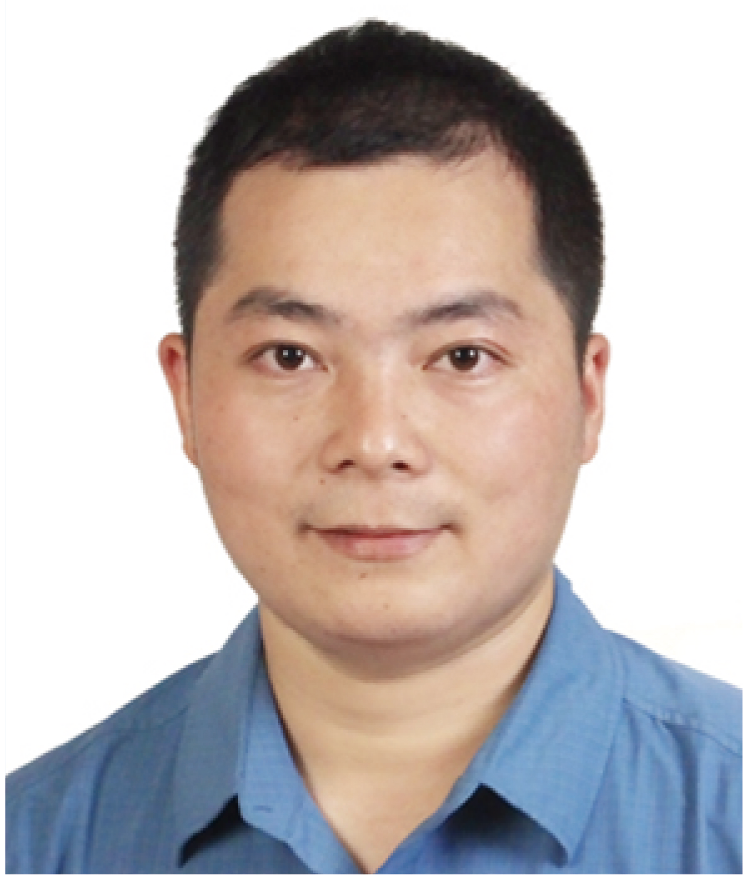}}]{Jing Zhang}
is a research fellow at the School of Computer Science of the University of Sydney. His research interests include computer vision and deep learning. He has published more than 30 papers in prestigious journals and proceedings at leading conferences. He serves as a reviewer for many journals and conferences and a senior program committee member of the AAAI Conference on Artificial Intelligence and the International Joint Conference on Artificial Intelligence. He is a member of IEEE and ACM.
\end{IEEEbiography}
\vspace{-12mm}
\begin{IEEEbiography}[{\includegraphics[width=0.9in,clip,keepaspectratio]{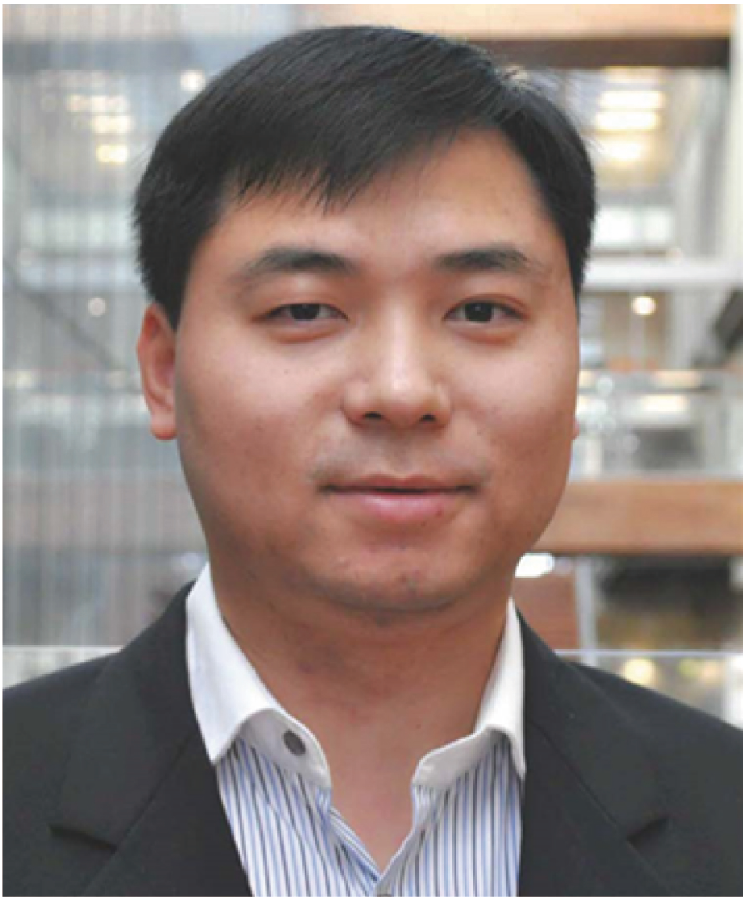}}]{Dacheng Tao (Fellow, IEEE)}
is the President of the JD Explore Academy and a Senior Vice President of JD.com. He is also an advisor and chief scientist of the digital science institute in the University of Sydney. He mainly applies statistics and mathematics to artificial intelligence and data science, and his research is detailed in one monograph and over 200 publications in prestigious journals and proceedings at leading conferences. He received the 2015 Australian Scopus-Eureka Prize, the 2018 IEEE ICDM Research Contributions Award, and the 2021 IEEE Computer Society McCluskey Technical Achievement Award. He is a fellow of the Australian Academy of Science, AAAS, ACM and IEEE.
\end{IEEEbiography}

\end{document}